\documentclass[10pt,journal,twoside,twocolumn]{IEEEtran}

\usepackage[utf8]{inputenc}
\usepackage{graphicx}
\usepackage{subfigure}
\usepackage{amsmath}
\usepackage{booktabs}

\begin{document}
	
	\title{Spatio-Temporal Dual Affine Differential Invariant for Skeleton-based Action Recognition}
	
	\author{Qi~Li,
		Hanlin~Mo,
		Jinghan~Zhao, 
		Hongxiang Hao,  
		and~Hua~Li,~\IEEEmembership{Senior Member,~IEEE}
		\thanks{This work is supported by National Key R\&D Program of China (No. 2019YFF0301801, No. 2017YFB1002703), 
			National Key Basic Research Program (No. 2015CB554507) 
			and National Natural Science Foundation of China under Grant 61379082.}
		\thanks{Qi Li is with the Key Laboratory of Intelligent Information Processing,
			Institute of Computing Technology, Chinese Academy of Science and University of Chinese Academy of Sciences (email:~liqi@ict.ac.cn).}
		\thanks{Hanlin Mo, Jinghan Zhao and Hongxiang Hao are with the Key Laboratory of Intelligent Information Processing, Institute of Computing Technology, Chinese Academy of Science and University of Chinese Academy of Sciences	(email:~mohanlin@ict.ac.cn, ~zhaojinghan17s@ict.ac.cn and~haohongxiang18s@ict.ac.cn).}
		\thanks{Hua Li is with the Key Laboratory of Intelligent Information Processing, Institute of Computing Technology, Chinese Academy of Science (email:~lihua@ict.ac.cn).}
	}
		
		
	\maketitle
		
	\begin{abstract}
		
		The dynamics of human skeletons have significant information for the task of action recognition. 
		The similarity between trajectories of corresponding joints is an indicating feature of the same action, while this similarity may subject to some distortions that can be modeled as the combination of spatial and temporal affine transformations.
		In this work, we propose a novel feature called spatio-temporal dual affine differential invariant (STDADI). 
		Furthermore, in order to improve the generalization ability of neural networks, a channel augmentation method is proposed. 
		On the large scale action recognition dataset NTU-RGB+D, and its extended version NTU-RGB+D 120, it achieves remarkable improvements over previous state-of-the-art methods. 
		
	\end{abstract}
		
	\section{Introduction}
		
		Skeleton-based action recognition has received great attention in recent years, as the dynamics of human skeletons has significant information for the task of action recognition. 
		Compared to other modalities of human action, for example, video, depth images and optical flows, human skeletons have the advantage of small amount of data and high information density. 
		The dynamics of human skeletons can be seen as time series of human poses, or the combination of human joint trajectories. 
		Among all the human joints, the trajectory of important joints indicating the action class conveys the most significant information. 
		It is also worth noting that when performing the same action under different attempts, trajectories of these joints are subject to some distortions. 
		In this work, we propose a novel invariant feature under these distortions and then utilize them for facilitating skeleton-based action recognition. 
		
		When performing the same action, two similar trajectories of corresponding joints should share a basic shape. 
		However, due to individual factors, these two trajectories always appear in diverse kinds of distortions. 
		These distortions are caused by spatial and temporal factors.
		Spatial factors include the change of viewpoints, different skeleton sizes and action amplitude~(\cite{wu2009flexible, wang2017modeling}), while temporal factors indicate time scaling along the time series~(\cite{esling2012time, he2014new}). 
		All the spatial factors can be modeled by the affine transformation in 3D space, whereas the uniform time scaling is commonly discussed case, which can be seen as affine transformation in 1D space. 
		We combine these two kinds of distortions as the spatio-temporal dual affine transformation. 
		
		In this paper, we propose a general method for constructing Spatio-Temporal Dual Affine Differential Invariant (STDADI). 
		Specifically, we utilize the rational polynomial of derivatives of joint trajectories to obtain the invariants. 
		By bounding the degree of polynomial and the order of derivatives, we generate 8 independent STDADIs and combine them as an invariant vector at each moment for each human joint. 
		
		Recently, researchers tend to explore the potential of date-driven methods for skeleton-based action recognition. 
		When considering to improve the generalization ability of neural networks under different transformations, a common practice is data augmentation. 
		However, additional data preprocessing generates more samples and takes longer time in the training phase. 
		In this paper, we propose an intuitive yet effective method, extending input data with STDADI along the channel dimension for training and evaluation, and call this practice as channel augmentation. 
		Experiments show that channel augmentation based on STDADI not only achieves stronger performance and generalization, but also provides more insights for skeleton-based action recognition. 
		
		The main contributions of this work are the following: 
		\begin{enumerate}
			\item We propose a novel feature called spatio-temporal dual affine differential invariant (STDADI). 
			\item In order to improve the generalization ability of neural networks, a channel augmentation method is proposed.
			\item We validate the effectiveness of the proposed feature and method on the large scale action recognition dataset NTU-RGB+D~\cite{shahroudy2016ntu} and its extended version NTU-RGB+D 120~\cite{liu2019ntu}, and get superior performance when compared to previous state-of-the-art methods.  
		\end{enumerate}
		
	\section{Related Work}
	
		{\bf Skeleton-based action recognition} 
		Before the rising of deep learning, some handcrafted-feature-based methods were proposed to solve skeleton-based action recogniton.
		Wang \MakeLowercase{\textit{et al.}}\cite{wang2012mining} proposed to use relative locations of joints as motion features. 
		Hussein \MakeLowercase{\textit{et al.}}\cite{hussein2013human} exploited the covariance matrices of joint trajectories. 
		Vemulapalli \MakeLowercase{\textit{et al.}}\cite{vemulapalli2014human} utilized rotations and translations between joint locations to capture the dynamics of human skeletons. 
		However, the performance of these methods is limited as designed features do not cover all factors affecting the recognition.
		Thanks to the success of deep learning, data-driven methods achieve better performance than before. 
		These methods can be further divided as RNN-based, CNN-based and GNN-based approaches.
		RNN-based approaches take the sequence of human joint coordinate vectors as input and predict the action label in a recursive manner~(\cite{wang2017modeling,shahroudy2016ntu,veeriah2015differential,du2015hierarchical,liu2016spatio,zhang2017view,liu2017global,liu2017skeleton}). 
		CNN-based approaches express the skeleton data as a pseudo-image for conventional CNNs~(\cite{liu2017enhanced,liu2017two,li2017skeleton,li2017skeleton2,liu2018recognizing}) or as a sequence of coordinate vectors for temporal CNNs~(\cite{ke2017new,kim2017interpretable,ke2018learning}). 
		Compared to these two kinds of methods, GNN-based approaches are modeled based on the natural connections between human joints, thus better to characterize the dynamics of human skeletons. 
		Recently, Yan \MakeLowercase{\textit{et al.}}\cite{yan2018spatial} proposed the spatiao-temporal graph convolutional network (ST-GCN) and achieved evident improvements over previous methods. 
	
		{\bf Transformations and invariant features for skeleton-based action recognition}
		For skeleton-based action recognition, commonly discussed transformations are geometric, which are usually caused by the change of viewpoint and the magnitude of motion.
		M{\"u}ller \MakeLowercase{\textit{et al.}}~\cite{muller2005efficient} took a set of boolean features associated with four joints to describe their relative positions, which is invariant with respect to the skeleton's position, orientation and size. 
		Vemulapalli \MakeLowercase{\textit{et al.}}~\cite{vemulapalli2014human} utilized rigid transformations between human joints to describe the skeleton, which are geometrically invariant. 
		Shao \MakeLowercase{\textit{et al.}}~\cite{shao2015integral} used integral invariants as a local description of joint trajectory.
		Boulahia \MakeLowercase{\textit{et al.}}~\cite{boulahia2016hif3d} integrated a set of features inspired by the handwriting recogniton, in which moment invariants~\cite{sadjadi1980three} with respect to similar transformation were utilized.
		
		Time scaling, as a transformation along the time dimension, are hardly discussed by previous works for skeleton-based action recognition.
		Most of these works(\cite{wu2009flexible,shao2015integral,anirudh2016elastic,kacem2018novel}), use a dynamic time warping technique for time alignment and trajectory matching.
		Esling and Agon~\cite{esling2012time} explicitly defined time scaling, and classfy it as uniform and dynamic.
		We explore the definition of uniform time scaling and model it as the temporal affine transformation.
		
		In the deep learning domain, data augmentation is a universal approach for improving generalization under various transformations.
		However, this method is time-consuming during training and hard to explain for its improvement. 
		Wang \MakeLowercase{\textit{et al.}}\cite{wang2017modeling} proposed a data augmentation method for 3D coordinates of human skeletons including rotation, scaling and shear transformations, and this method is beneficial to training of the proposed two-stream RNN.
		
	\section{Approach}
	
		\subsection{Spatio-temporal Dual Affine Transformation} 
		
		Formally, we express the dynamics of human joints in the form of parameterized curve taking time as the parameter:
		\begin{align}
		\mathbf{f}(t) & = \left( f_x(t), f_y(t), f_z(t) \right)^T \\  
		\mathbf{g}(u) & = \left( g_x(u), g_y(u), g_z(u) \right)^T
		\end{align} 
		where $\mathbf{f}(t)$ and $\mathbf{g}(u)$ represent joint trajectories before and after the transformation respectively.
		
		The dual affine transformation can be defined as  
		\begin{equation}
		\mathbf{g}(u)  = \mathbf{A} \mathbf{f}(t) + \mathbf{T}, u  = \frac{1}{c} (t-d)
		\label{Eq:transformation}
		\end{equation}
		where the matrix $\mathbf{A}$ and vector $\mathbf{T}$ express the spatial affine transformation, and the scalar $c$ and $d$ are used to denote the temporal affine transformation. This can be detailed as follows:
		 
		{\bf Spatial affine transformation}
		The matrix $\mathbf{A}$ controls the rotation and scaling while the vector $\mathbf{T}$ means the translation. 
		Spatial affine transformations are caused by multiple factors, including coordinate system convertion, pose orientation, different skeleton sizes and action amplitude.  
		
		{\bf Temporal affine transformation}
		The linear transformation of time domain can be considered as the 1D affine transformation. 
		The parameter $c$ means time scaling, indicating different speeds, and $d$ means phase shift, indicating different beginning time. 
		We discuss uniform time scaling here and it assumes a uniform change of the time scale according to the same proportion~\cite{he2014new}.
		We follow this assumption and express it as the temporal affine transformation.
		
		\subsection{Spatio-Temporal Dual Affine Differential Invariant}  
		
		We utilize the rational polynomial of derivatives of joint trajectories to construct STDADI. 
		Specifically, based on equation~\ref{Eq:transformation}, we can derive the relationship between 1st derivatives of joint trajectories before and after the transformation: 
		\begin{align}
		\frac{d\mathbf{g}}{du} & = \frac{d\mathbf{g}}{d\mathbf{f}}\cdot \frac{d\mathbf{f}}{dt}\cdot \frac{dt}{du} \\ \notag
		& = \mathbf{A} \cdot \frac{d\mathbf{f}}{dt} \cdot c 
		  = c\mathbf{A} \cdot \frac{d\mathbf{f}}{dt} 
		\end{align}
		
		Similarly, we can obtain the relationship between their any order derivatives by chain rule: 
		\begin{equation}
		\mathbf{g}^{(i)} = c^i\mathbf{A}\mathbf{f}^{(i)} 
		\label{Eq:derivative}
		\end{equation}
		where the superscript $(i)$ denotes the order of derivation. 
		
		It is worth noting that when $i$ is equal to 0, formula~\ref{Eq:derivative} is equivalent to formula~\ref{Eq:transformation} without translation vector $T$. 
		We can eliminate the effect of translation vector $\mathbf{T}$ by subtracting the mean value. 
		That is, 
		\begin{equation}
		\mathbf{\hat{g}} = \mathbf{g}-\mathbf{g}_{mean} = \mathbf{A}\cdot (\mathbf{f}-\mathbf{f}_{mean}) = c^0\mathbf{A}\mathbf{\hat{f}}
		\end{equation}
		Thus, in Equation~\ref{Eq:derivative}, we can set $i$ as a non-negative integer. 
		
		Based on the relationship in equation~\ref{Eq:derivative}, we construct a $3\times3$ matrix using 3 derivatives of different orders as column and derive their relationship: 
		\begin{align}
		\mathbf{M}^{ijk}_g & = \left( \mathbf{g}^{(i)},\mathbf{g}^{(j)},\mathbf{g}^{(k)} \right) \\ \notag 
		& = \left( c^i\mathbf{A}\mathbf{f}^{(i)},c^j\mathbf{A}\mathbf{f}^{(j)},c^k\mathbf{A}\mathbf{f}^{(k)} \right) \\ \notag 
		& = \mathbf{A} \cdot \left( c^i\mathbf{f}^{(i)},c^j\mathbf{f}^{(j)},c^k\mathbf{f}^{(k)} \right) 
		\end{align}
		where $i,j,k$ are all non-negative integers. 
		To ensure the determinant of $\mathbf{M}$ is not equal to 0, $i,j,k$ are different from each other. 
		We find that the determinant of $\mathbf{M}$ is a relative invariant which is related to the transformation parameters of $c$ and $\mathbf{A}$: 
		\begin{align}
		\|\mathbf{M}^{ijk}_g\| & = c^{i+j+k} \|\mathbf{A}\| \|(\mathbf{f}^{(i)},\mathbf{f}^{(j)},\mathbf{f}^{(k)})\| \\ \notag 
		& = c^{i+j+k} \|\mathbf{A}\| \|\mathbf{M}^{ijk}_f\|
		\end{align} 
		
		We eliminate the parameters of $c$ and $\mathbf{A}$ by constructing the rational formula: 
		\begin{equation}
		\frac{\prod_{\sigma=1}^{N}\|\mathbf{M}^{i_{\sigma}j_{\sigma}k_{\sigma}}_g\|}{\prod_{\sigma=1}^{N}\|\mathbf{M}^{l_{\sigma}m_{\sigma}n_{\sigma}}_g\|} = \frac{\prod_{\sigma=1}^{N}\|\mathbf{M}^{i_{\sigma}j_{\sigma}k_{\sigma}}_f\|}{\prod_{\sigma=1}^{N}\|\mathbf{M}^{l_{\sigma}m_{\sigma}n_{\sigma}}_f\|}
		\end{equation}
		
		This means that 
		\begin{equation}
		\frac{\prod_{\sigma=1}^{N}\|\mathbf{M}^{i_{\sigma}j_{\sigma}k_{\sigma}}\|}{\prod_{\sigma=1}^{N}\|\mathbf{M}^{l_{\sigma}m_{\sigma}n_{\sigma}}\| + \epsilon}
		\label{Eq:invariant}
		\end{equation}
		is an invariant feature with respect to the spatio-temporal dual affine transformation, namely, STDADI. 
		In this expression, N is a positive integer named as the degree of polynomials, and the degree of numerator and denominator should be equal to guarantee the elimination of the matrix $\mathbf{A}$.
		The max value of derivatives is named as the order of STDADI. 
		To ensure the elimination of the parameter $c$, the following needs to be met, 
		\begin{equation}
		\sum_{\sigma=1}^{N}(i_{\sigma} + j_{\sigma} + k_{\sigma}) = \sum_{\sigma=1}^{N}(l_{\sigma} + m_{\sigma} + n_{\sigma})
		\end{equation}
		To ensure that every determinant is not equal to 0, it is also needed that
		$i_{\sigma} \ne j_{\sigma} \ne k_{\sigma}, l_{\sigma} \ne m_{\sigma} \ne n_{\sigma}, \forall \sigma$.
		The parameter $\epsilon$ is a small value for computational stability. 
		
		For computation simplicity, we set the upper limit of the degree and order to be 2 and 4, respectively, and we obtain 55 invariants in total.
		We select 8 of them which are function-independent~\cite{brown1935functional} from each other, which means weaker correlation and better description ability. 
		The 8 invariants are listed as follows ($\epsilon$ are ignored here for compact expression): 
		\begin{align}
		& \frac{\|\mathbf{M}^{023}\|}{\|\mathbf{M}^{014}\|}, \frac{\|\mathbf{M}^{123}\|}{\|\mathbf{M}^{024}\|}, 
		\frac{\|\mathbf{M}^{034}\|}{\|\mathbf{M}^{124}\|}, \label{Eq:function-independent} \\ \notag 
		& \frac{\|\mathbf{M}^{012}\|\|\mathbf{M}^{023}\|}{\|\mathbf{M}^{013}\|^2}, 
		\frac{\|\mathbf{M}^{013}\|\|\mathbf{M}^{123}\|}{\|\mathbf{M}^{014}\|^2},
		\frac{\|\mathbf{M}^{023}\|\|\mathbf{M}^{124}\|}{\|\mathbf{M}^{123}\|^2}, \\ \notag
		& \frac{\|\mathbf{M}^{123}\|\|\mathbf{M}^{134}\|}{\|\mathbf{M}^{124}\|^2},
		\frac{\|\mathbf{M}^{124}\|\|\mathbf{M}^{234}\|}{\|\mathbf{M}^{134}\|^2}
		\end{align}
		
		In practice, we approxiamate the derivatives of joint trajectories using a 5th order B-spline curve. 
		Then we calculate STDADIs following formula~\ref{Eq:invariant} and~\ref{Eq:function-independent}. 
		Finally we arrange the obtained invariants as an 8-dimension invariant feature vector at each moment for each human joint.  
		
		\subsection{Channel Augmentation} 
		
		Compared to other handcrafted features, our STDADI focuses on describing joint trajectories under the spatio-temporal dual affine transformation.
		As not all factors are covered, STDADI itself is not efficient enough for the recognition task. 
		However, as the feature is beneficial for recognizing actions under different transformations, it can help improve the generalization of data-driven methods.
		In this case, we propose an inituitive yet effective method named channel augmentation.
		
\begin{figure}	
			\centering
			\includegraphics[width=0.65\linewidth]{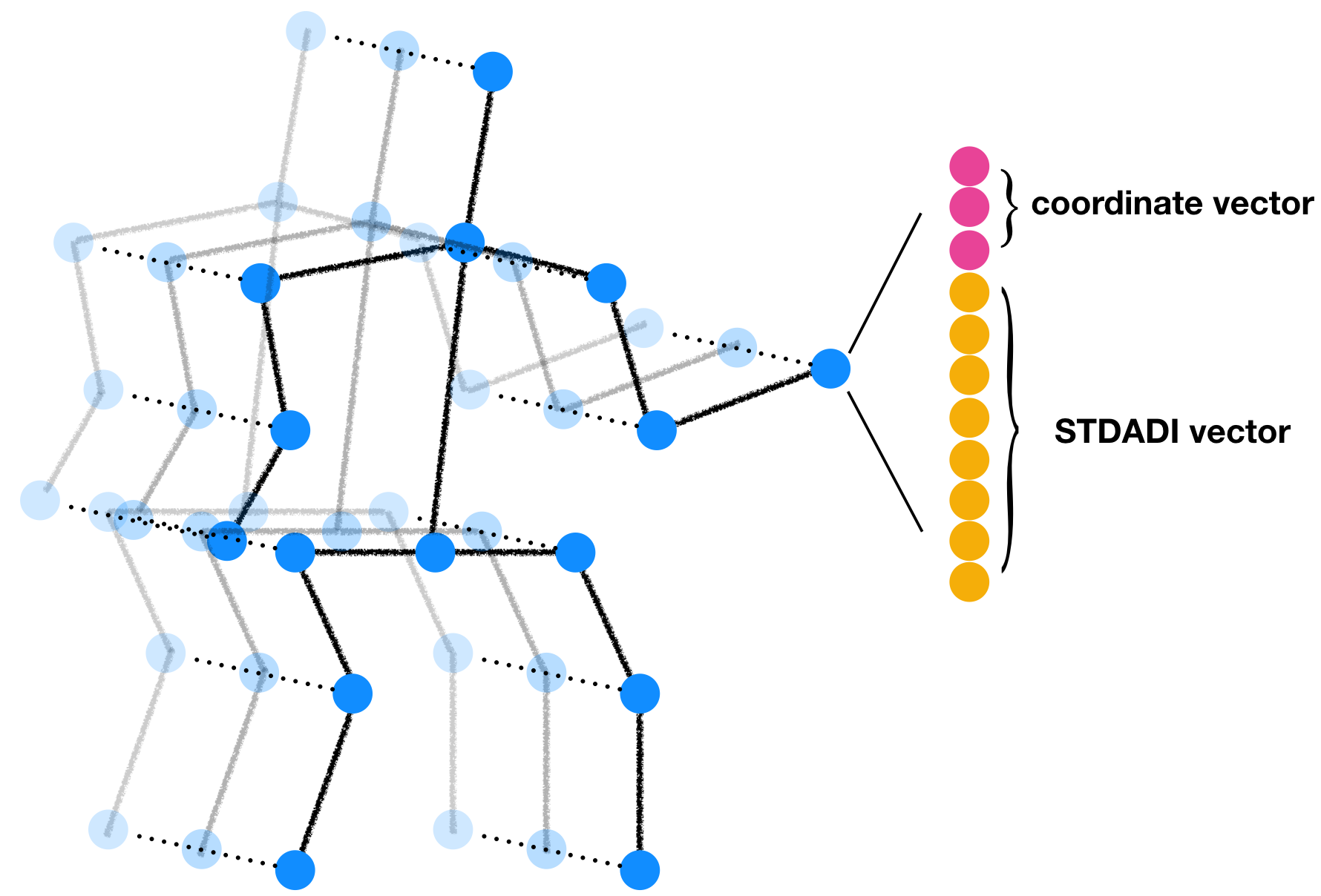}
			\caption{Illustration of channel augmentation for the spatio-temporal graph of human skeletons.}
			\label{fig:channelAug}	
		\end{figure}

		Specifically, we extend input data with STDADI along the channel dimension, as shown in Fig.~\ref{fig:channelAug}. 
		Conventional inputs are 3D coordinates of human joints, and we concatenate the coordinate vector and the STDADI vector at each joint for each frame. 
		Before the concatenation, we apply a hyperbolic tangent function on the STDADI vector to make sure that it matches the magnitude of coordinates.
		Channel augmentation introduces invariant information into input data without changing the inner structure of neural networks.
		
		In our experiments we choose to use spatio-temporal graph convolutional networks (ST-GCN)~\cite{yan2018spatial}.
		This method models the skeleton data as a graph structure, considering spatial and temporal connections between human joints simultaneously.
		Particularly, it can help exploit local pattern and correlation from human skeletons, in other words, the importance of joints along the action sequence, expressed as weights of joints in the spatio-temporal graph. 
		This is in line with our STDADI, because both of them focus on describing joint dynamics, and our features further provide an invariant expression which is not affected by the distortions.
		
				\begin{table*}
			\centering
			\normalsize
			\caption{Comparisons of the validation accuracy with state-of-the-art methods on NTU-RGB+D and NTU-RGB+D 120. "*" indicates that for~\cite{liu2018recognizing}, we report here the results on NTU-RGB+D using only skeleton data. Best results are labeled in bold.}
			\label{table:ntu}
			\scalebox{0.8}{
				\begin{tabular}{lccccc}
					\toprule
					& & \multicolumn{2}{c}{NTU-RGB+D} & \multicolumn{2}{c}{NTU-RGB+D 120} \\
					Method & & Cross-subject & Cross-view & Cross-subject & Cross-setup \\
					\midrule
					Part-Aware LSTM & \cite{shahroudy2016ntu} & 62.9\% & 70.3\% & 25.5\% & 26.3\% \\
					Spatio-Termporal LSTM & \cite{liu2016spatio} & 69.2\% & 77.7\% & 55.7\% & 57.9\% \\ 
					GCA-LSTM & \cite{liu2017global} & 74.4\% & 82.8\% & 58.3\% & 59.2\% \\ 
					Two-Stream Attention LSTM & \cite{liu2017skeleton} & 76.1\% & 84.0\% & 61.2\% & 63.3\% \\
					Skeleton Visualization & \cite{liu2017enhanced} & 80.0\% & 87.2\% & 60.3\% & 63.2\% \\
					Body Pose Evolution Map(*) & \cite{liu2018recognizing} & 82.4\% & 86.7\% & 64.6\% & 66.9\% \\	
					Multi-Task Learning Network & \cite{ke2017new} & 79.6\% & 84.8\% & 58.4\% & 57.9\% \\ 
					Multi-Task CNN with RotClips & \cite{ke2018learning} & 81.1\% & 87.4\% & 62.2\% & 61.8\% \\
					\hline
					ST-GCN & \cite{yan2018spatial} & 81.5\% & 88.3\% & 71.7\% & 74.3\% \\ 
					ST-GCN + data augmentation & & 80.6\% & 90.5\% & 72.2\% & {\bf 79.0\%} \\ 
					ST-GCN + channel augmentation & & {\bf 83.4\%} & {\bf 91.3\%} & {\bf 77.3\%} & 78.8\% \\
					\bottomrule 
				\end{tabular}
			}
		\end{table*}
		
	\section{Results}
	
		In this section we validate the effectiveness of the proposed feature and method on the large scale action recognition dataset NTU-RGB+D~\cite{shahroudy2016ntu} and its extended version NTU-RGB+D 120~\cite{liu2019ntu}. 
		In addition to the original ST-GCN, we adopted a data augmentation technique as the baseline method.
		As illustrated in~\cite{wang2017modeling}, the data augmentation technique involves rotation, scaling and shear transformations of 3D skeletons during training.
		For all the experimental methods, we used the same training strategy and hyperparameters as suggested in~\cite{yan2018spatial}. 
	
		\subsection{Datasets \& Evaluation Metrics} 
		
		NTU-RGB+D and its extended version, NTU-RGB+D 120 are currently the largest action recognition datasets with 3D joint annotations captured in a constrained indoor environment using Microsoft Kinect V2 cameras.
		Both of them provide 3D skeleton data containing 3D locations of 25 major body joints in the camera coordinate system.
		NTU-RGB+D contains 56880 samples in 60 action classes performed by 40 subjects, and NTU RGB+D 120 extends the original by adding 57600 more samples, expanding the number of action classes and subjects to 120 and 106, respectively.
		Both datasets have the cross-subject evaluation criteria, while NTU RGB+D 120 makes an improvement on the cross-view benchmark by introducing more factors that affect the angle of view, including the height and distance of cameras to the subjects, and renames this benchmark as "cross-setup". 
		We report top-1 recognition accuracy on both datasets with corresponding evaluation metrics. 	
	
		\subsection{Comparison with the State-of-the-art} 
		
		As shown in Table~\ref{table:ntu}, our method, ST-GCN + channel augmentation, outperforms most of the previous state-of-the-art methods. 
		Compare to two baseline approaches, ST-GCN and ST-GCN + data augmentation, our method achieves obvious improvements on both benchmarks.
		For data augmentation, as it is mainly consisted of 3D geometric transformations, it helps much to improve accuracy in cross-view recognition, but contributes little to the cross-subject setting.
		This also verifies that our spatio-temporal dual affine transformation assumption is valid on both evaluation criteria.
	
		\begin{table}
			\centering
			\normalsize
			\caption{The validation accuracy of different input settings for channel augmentation on NTU-RGB+D.}
			\label{table:input}
			\scalebox{0.8}{
				\begin{tabular}{ccc}
					\toprule
					Method & Cross-subject & Cross-view \\
					\midrule
					ST-GCN & 81.5\% & 88.3\% \\ 
					+ derivatives & 80.4\% & 87.6\% \\ 
					+ STDADI & {\bf 83.4\%} & {\bf 91.3\%} \\
					\bottomrule
				\end{tabular}
			}
		\end{table}	
	
		\begin{figure}
			\begin{center}
				\includegraphics[width=0.95\linewidth]{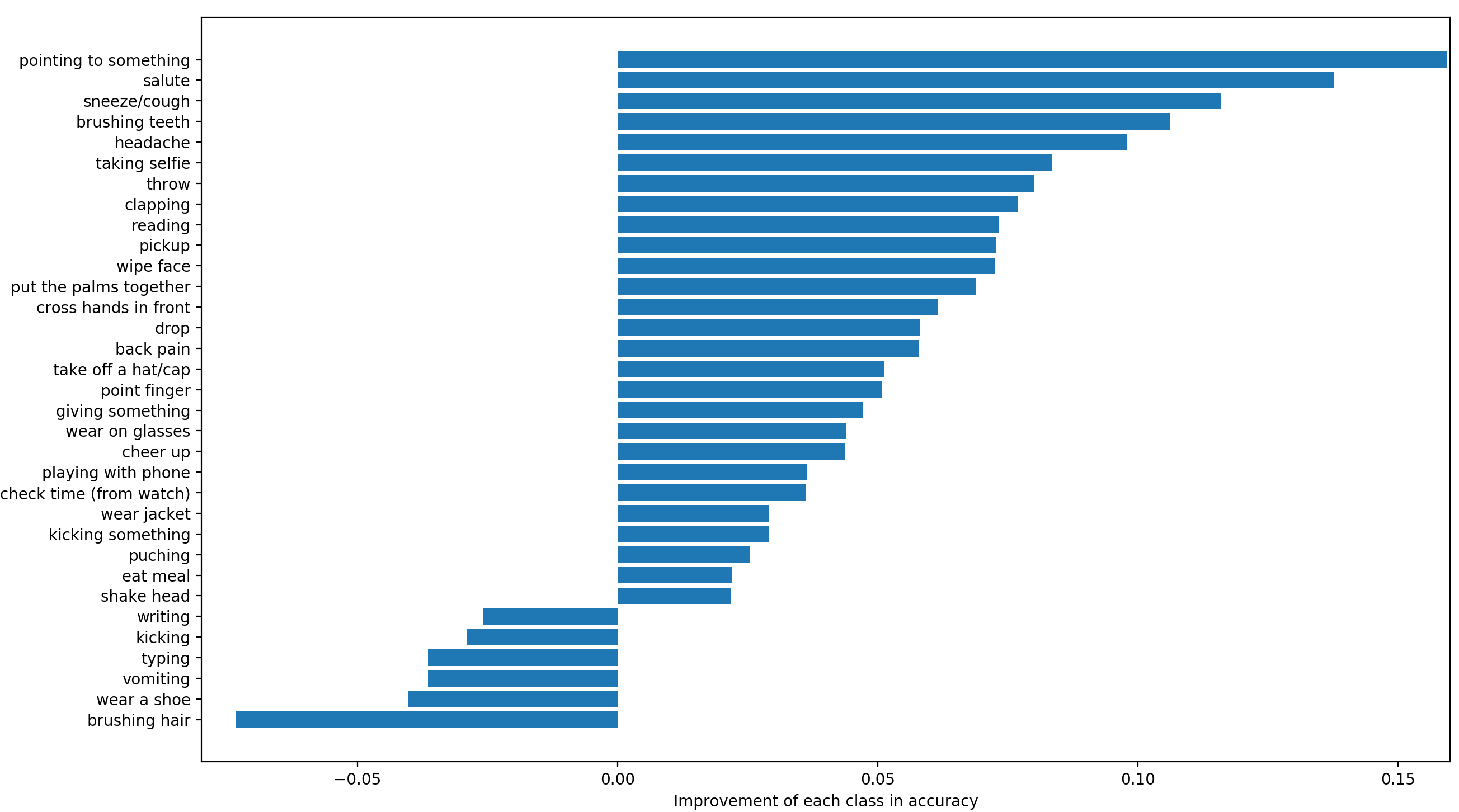}
			\end{center}
			\caption{Improvements of each class in validation accuracy of ST-GCN + channel augmentation over ST-GCN, on NTU-RGB+D, the cross-subject benchmark. For clarity only classes with accuracy improvement larger than 2\% or less than -2\% are shown.}
			\label{fig:accuracy}
		\end{figure}
		
		\subsection{Detailed Analysis}
		
		To validate the effectiveness of STDADI, we tried a different input setting using trajectory derivatives as the extended vector for channel augmentation. 
		This vector contains the 1st, 2nd and 3rd derivatives of the joint trajectory and thus is 9-dimensional. 
		Seen from Table~\ref{table:input}, while the ST-GCN+STDADI has an improvement, the ST-GCN+derivatives has a decrease of accuracy on both the benchmarks. 
		This shows that the improvement of accuracy comes from the invariance expressed by STDADI. 
		
		We also compare the improvements of ST-GCN + channel augmentation over ST-GCN of different action classes. 
		As shown in Fig.~\ref{fig:accuracy}, actions such as {\em pointing to something} and {\em salute} achieve the greatest performance gain, while actions like {\em brushing hairs} suffer performance loss. 
		We find that those action classes with improving accuracy have specific joint trajectory motion patterns. 
		When performing actions like {\em pointing to something} and {\em salute}, the trajectories of wrist joint of performers are geometrically similar. 
		This indicates that the geometric similarity of important joint trajectories helps to recognize the action class, and our STDADI provides an invariant representation for the similarity under various distortions.
	
	\section{Conclusion}
	
		In this paper, we propose a general method for constructing spatio-temporal dual affine differential invariant (STDADI). 
		We prove the effectiveness of this invariant feature using a channel augmentation technique on the large-scale action recognition dataset NTU-RGB+D and NTU-RGB+D 120. 
		The combination of handcrafted features and data-driven methods not only improves the accuracy but also provides more insights. 
		In the future, as the temporal affine transformation may not be efficient to model complex transformations along the time dimension, we are going to explore the invariance under unlinear dynamic time scaling. 
	
		\bibliographystyle{IEEEtran}
		\bibliography{STDADI}

\begin{thebibliography}{10}
\providecommand{\url}[1]{#1}
\csname url@samestyle\endcsname
\providecommand{\newblock}{\relax}
\providecommand{\bibinfo}[2]{#2}
\providecommand{\BIBentrySTDinterwordspacing}{\spaceskip=0pt\relax}
\providecommand{\BIBentryALTinterwordstretchfactor}{4}
\providecommand{\BIBentryALTinterwordspacing}{\spaceskip=\fontdimen2\font plus
\BIBentryALTinterwordstretchfactor\fontdimen3\font minus
  \fontdimen4\font\relax}
\providecommand{\BIBforeignlanguage}[2]{{%
\expandafter\ifx\csname l@#1\endcsname\relax
\typeout{** WARNING: IEEEtran.bst: No hyphenation pattern has been}%
\typeout{** loaded for the language `#1'. Using the pattern for}%
\typeout{** the default language instead.}%
\else
\language=\csname l@#1\endcsname
\fi
#2}}
\providecommand{\BIBdecl}{\relax}
\BIBdecl

\bibitem{wu2009flexible}
S.~Wu and Y.~F. Li, ``Flexible signature descriptions for adaptive motion
  trajectory representation, perception and recognition,'' \emph{Pattern
  Recognition}, vol.~42, no.~1, pp. 194--214, 2009.

\bibitem{wang2017modeling}
H.~Wang and L.~Wang, ``Modeling temporal dynamics and spatial configurations of
  actions using two-stream recurrent neural networks,'' in \emph{Proceedings of
  the IEEE Conference on Computer Vision and Pattern Recognition}, 2017, pp.
  499--508.

\bibitem{esling2012time}
P.~Esling and C.~Agon, ``Time-series data mining,'' \emph{ACM Computing Surveys
  (CSUR)}, vol.~45, no.~1, p.~12, 2012.

\bibitem{he2014new}
X.~He, C.~Shao, and Y.~Xiong, ``A new similarity measure based on shape
  information for invariant with multiple distortions,'' \emph{Neurocomputing},
  vol. 129, pp. 556--569, 2014.

\bibitem{shahroudy2016ntu}
A.~Shahroudy, J.~Liu, T.-T. Ng, and G.~Wang, ``Ntu rgb+ d: A large scale
  dataset for 3d human activity analysis,'' in \emph{Proceedings of the IEEE
  conference on computer vision and pattern recognition}, 2016, pp. 1010--1019.

\bibitem{liu2019ntu}
J.~Liu, A.~Shahroudy, M.~L. Perez, G.~Wang, L.-Y. Duan, and A.~K. Chichung,
  ``Ntu rgb+ d 120: A large-scale benchmark for 3d human activity
  understanding,'' \emph{IEEE transactions on pattern analysis and machine
  intelligence}, 2019.

\bibitem{wang2012mining}
J.~Wang, Z.~Liu, Y.~Wu, and J.~Yuan, ``Mining actionlet ensemble for action
  recognition with depth cameras,'' in \emph{2012 IEEE Conference on Computer
  Vision and Pattern Recognition}.\hskip 1em plus 0.5em minus 0.4em\relax IEEE,
  2012, pp. 1290--1297.

\bibitem{hussein2013human}
M.~E. Hussein, M.~Torki, M.~A. Gowayyed, and M.~El-Saban, ``Human action
  recognition using a temporal hierarchy of covariance descriptors on 3d joint
  locations,'' in \emph{Twenty-Third International Joint Conference on
  Artificial Intelligence}, 2013.

\bibitem{vemulapalli2014human}
R.~Vemulapalli, F.~Arrate, and R.~Chellappa, ``Human action recognition by
  representing 3d skeletons as points in a lie group,'' in \emph{Proceedings of
  the IEEE conference on computer vision and pattern recognition}, 2014, pp.
  588--595.

\bibitem{veeriah2015differential}
V.~Veeriah, N.~Zhuang, and G.-J. Qi, ``Differential recurrent neural networks
  for action recognition,'' in \emph{Proceedings of the IEEE international
  conference on computer vision}, 2015, pp. 4041--4049.

\bibitem{du2015hierarchical}
Y.~Du, W.~Wang, and L.~Wang, ``Hierarchical recurrent neural network for
  skeleton based action recognition,'' in \emph{Proceedings of the IEEE
  conference on computer vision and pattern recognition}, 2015, pp. 1110--1118.

\bibitem{liu2016spatio}
J.~Liu, A.~Shahroudy, D.~Xu, and G.~Wang, ``Spatio-temporal lstm with trust
  gates for 3d human action recognition,'' in \emph{European Conference on
  Computer Vision}.\hskip 1em plus 0.5em minus 0.4em\relax Springer, 2016, pp.
  816--833.

\bibitem{zhang2017view}
P.~Zhang, C.~Lan, J.~Xing, W.~Zeng, J.~Xue, and N.~Zheng, ``View adaptive
  recurrent neural networks for high performance human action recognition from
  skeleton data,'' in \emph{Proceedings of the IEEE International Conference on
  Computer Vision}, 2017, pp. 2117--2126.

\bibitem{liu2017global}
J.~Liu, G.~Wang, P.~Hu, L.-Y. Duan, and A.~C. Kot, ``Global context-aware
  attention lstm networks for 3d action recognition,'' in \emph{Proceedings of
  the IEEE Conference on Computer Vision and Pattern Recognition}, 2017, pp.
  1647--1656.

\bibitem{liu2017skeleton}
J.~Liu, G.~Wang, L.-Y. Duan, K.~Abdiyeva, and A.~C. Kot, ``Skeleton-based human
  action recognition with global context-aware attention lstm networks,''
  \emph{IEEE Transactions on Image Processing}, vol.~27, no.~4, pp. 1586--1599,
  2017.

\bibitem{liu2017enhanced}
M.~Liu, H.~Liu, and C.~Chen, ``Enhanced skeleton visualization for view
  invariant human action recognition,'' \emph{Pattern Recognition}, vol.~68,
  pp. 346--362, 2017.

\bibitem{liu2017two}
H.~Liu, J.~Tu, and M.~Liu, ``Two-stream 3d convolutional neural network for
  skeleton-based action recognition,'' \emph{arXiv preprint arXiv:1705.08106},
  2017.

\bibitem{li2017skeleton}
C.~Li, Q.~Zhong, D.~Xie, and S.~Pu, ``Skeleton-based action recognition with
  convolutional neural networks,'' in \emph{2017 IEEE International Conference
  on Multimedia \& Expo Workshops (ICMEW)}.\hskip 1em plus 0.5em minus
  0.4em\relax IEEE, 2017, pp. 597--600.

\bibitem{li2017skeleton2}
B.~Li, Y.~Dai, X.~Cheng, H.~Chen, Y.~Lin, and M.~He, ``Skeleton based action
  recognition using translation-scale invariant image mapping and multi-scale
  deep cnn,'' in \emph{2017 IEEE International Conference on Multimedia \& Expo
  Workshops (ICMEW)}.\hskip 1em plus 0.5em minus 0.4em\relax IEEE, 2017, pp.
  601--604.

\bibitem{liu2018recognizing}
M.~Liu and J.~Yuan, ``Recognizing human actions as the evolution of pose
  estimation maps,'' in \emph{Proceedings of the IEEE Conference on Computer
  Vision and Pattern Recognition}, 2018, pp. 1159--1168.

\bibitem{ke2017new}
Q.~Ke, M.~Bennamoun, S.~An, F.~Sohel, and F.~Boussaid, ``A new representation
  of skeleton sequences for 3d action recognition,'' in \emph{Proceedings of
  the IEEE conference on computer vision and pattern recognition}, 2017, pp.
  3288--3297.

\bibitem{kim2017interpretable}
T.~S. Kim and A.~Reiter, ``Interpretable 3d human action analysis with temporal
  convolutional networks,'' in \emph{2017 IEEE Conference on Computer Vision
  and Pattern Recognition Workshops (CVPRW)}.\hskip 1em plus 0.5em minus
  0.4em\relax IEEE, 2017, pp. 1623--1631.

\bibitem{ke2018learning}
Q.~Ke, M.~Bennamoun, S.~An, F.~Sohel, and F.~Boussaid, ``Learning clip
  representations for skeleton-based 3d action recognition,'' \emph{IEEE
  Transactions on Image Processing}, vol.~27, no.~6, pp. 2842--2855, 2018.

\bibitem{yan2018spatial}
S.~Yan, Y.~Xiong, and D.~Lin, ``Spatial temporal graph convolutional networks
  for skeleton-based action recognition,'' in \emph{Thirty-Second AAAI
  Conference on Artificial Intelligence}, 2018.

\bibitem{muller2005efficient}
M.~M{\"u}ller, T.~R{\"o}der, and M.~Clausen, ``Efficient content-based
  retrieval of motion capture data,'' in \emph{ACM Transactions on Graphics
  (ToG)}, vol.~24, no.~3.\hskip 1em plus 0.5em minus 0.4em\relax ACM, 2005, pp.
  677--685.

\bibitem{shao2015integral}
Z.~Shao and Y.~Li, ``Integral invariants for space motion trajectory matching
  and recognition,'' \emph{Pattern Recognition}, vol.~48, no.~8, pp.
  2418--2432, 2015.

\bibitem{boulahia2016hif3d}
S.~Y. Boulahia, E.~Anquetil, R.~Kulpa, and F.~Multon, ``Hif3d:
  Handwriting-inspired features for 3d skeleton-based action recognition,'' in
  \emph{2016 23rd International Conference on Pattern Recognition
  (ICPR)}.\hskip 1em plus 0.5em minus 0.4em\relax IEEE, 2016, pp. 985--990.

\bibitem{sadjadi1980three}
F.~A. Sadjadi and E.~L. Hall, ``Three-dimensional moment invariants,''
  \emph{IEEE Transactions on Pattern Analysis and Machine Intelligence}, no.~2,
  pp. 127--136, 1980.

\bibitem{anirudh2016elastic}
R.~Anirudh, P.~Turaga, J.~Su, and A.~Srivastava, ``Elastic functional coding of
  riemannian trajectories,'' \emph{IEEE transactions on pattern analysis and
  machine intelligence}, vol.~39, no.~5, pp. 922--936, 2016.

\bibitem{kacem2018novel}
A.~Kacem, M.~Daoudi, B.~B. Amor, S.~Berretti, and J.~C. Alvarez-Paiva, ``A
  novel geometric framework on gram matrix trajectories for human behavior
  understanding,'' \emph{IEEE transactions on pattern analysis and machine
  intelligence}, 2018.

\bibitem{brown1935functional}
A.~B. Brown, ``Functional dependence,'' \emph{Transactions of the American
  Mathematical Society}, vol.~38, no.~2, pp. 379--394, 1935.

\end{thebibliography}
\end{document}